\documentclass[conference]{IEEEtran}
\IEEEoverridecommandlockouts
\usepackage{cite}
\usepackage{amsmath,amssymb,amsfonts}
\usepackage{algorithmic}
\usepackage{graphicx}
\usepackage{textcomp}
\usepackage{xcolor}
\usepackage{booktabs}
\usepackage{comment}
\usepackage{tabu} 
\usepackage{makecell}
\usepackage{pifont}
\usepackage{marvosym}
\usepackage{multirow}
\usepackage{url} 
\usepackage{authblk} 
\pdfoutput=1 
\def\BibTeX{{\rm B\kern-.05em{\sc i\kern-.025em b}\kern-.08em
    T\kern-.1667em\lower.7ex\hbox{E}\kern-.125emX}}
\begin{document}

\title{Semantic Scene Completion with Multi-Feature Data Balancing Network}

\author[1,2]{Mona Alawadh\thanks{e-mail: m.alawadh@soton.ac.uk}}
\author[1]{Mahesan Niranjan\thanks{e-mail: mn@ecs.soton.ac.uk}}
\author[1]{Hansung Kim\thanks{e-mail: H.Kim@soton.ac.uk}}

\affil[1]{University of Southampton}
\affil[2]{Imam Mohammad Ibn Saud Islamic University}

\maketitle

\begin{abstract}
Semantic Scene Completion (SSC) is a critical task in computer vision, that utilized in applications such as virtual reality (VR). SSC aims to construct detailed 3D models from partial views by transforming a single 2D image into a 3D representation, assigning each voxel a semantic label. The main challenge lies in completing 3D volumes with limited information, compounded by data imbalance, inter-class ambiguity, and intra-class diversity in indoor scenes. To address this, we propose the Multi-Feature Data Balancing Network (MDBNet), a dual-head model for RGB and depth data (F-TSDF) inputs. Our hybrid encoder-decoder architecture with identity transformation in a pre-activation residual module (ITRM) effectively manages diverse signals within F-TSDF. We evaluate RGB feature fusion strategies and use a combined loss function—cross-entropy for 2D RGB features and weighted cross-entropy for 3D SSC predictions. MDBNet results surpass comparable state-of-the-art (SOTA) methods on NYU datasets, demonstrating the effectiveness of our approach.
\end{abstract}

\vspace{-0.3cm}
\section{Introduction}
\label{sec:intro}
\vspace{-0.2cm}
Scene understanding is a fundamental aspect of computer vision, as it is essential for various real-world applications, including robotic navigation, virtual reality, and augmented reality \cite{alawadh2022room,gao2022review}.
Semantic Scene Completion (SSC) enhances these applications by providing comprehensive scene interpretations \cite{liang2021sscnav}. SSC task applied in VR applications such in \cite{alawadh2022room, kim2022immersive}. It aims to generate detailed and complete 3D models from partial views, typically utilizing depth maps and/or RGB images, by predicting the occupancy and semantic categories of objects within the scene. A notable example of SSC is SSCNet \cite{song2017semantic}, which integrates scene completion and semantic segmentation for indoor environments, illustrating the interdependence of these tasks and their mutual enhancement \cite{song2017semantic,roldao20223d}. Due to the partial-view nature of input data, SSC faces significant challenges, particularly the loss of 3D information in occluded regions. This makes predicting volumetric occupancy in these areas highly complex. Additionally, assigning accurate semantic labels within 3D spaces is complicated by factors such as dataset imbalance, intra-class diversity, and inter-class ambiguity \cite{pan2023understanding}. While some studies have addressed data imbalance through weighted loss functions, as seen in \cite{song2017semantic, zhang2019cascaded, li2019depth, tang2022not, dourado2021edgenet, dourado2022data}, they often overlook category imbalance within datasets. The work in \cite{alawadh20243d} tackled class imbalance by introducing a weighted cross-entropy function combined with a re-weighting method based on resampling and unsupervised clustering. Although this approach improved the recognition of certain classes, it struggled with challenging objects, such as windows and TVs. Windows often feature reflective or transparent surfaces, while TVs share visual characteristics with other categories, such as generic objects, making them difficult to distinguish using depth information alone in datasets like NYUv2 \cite{silberman2012indoor} and NYUCAD \cite{firman2016structured}. To tackle these challenges, we extend the method in \cite{alawadh20243d} by proposing a dual-head network with a combined loss function. Inspired by \cite{zhang2019cascaded, Park2019DeepSDFLC, Weder2020NeuralFusionOD} our approach incorporates the 3D Identity Transformed within full pre-activation Residual Module (ITRM), an innovative adaptation in the 3D CNN branch of MDBNet. This design introduces hyperbolic tangent activation (Tanh) on identity features, enabling effective processing of both positive and negative signals from F-TSDF inputs while normalizing feature distributions between -1 and 1.  Additionally, we explore various strategies for fusing RGB semantic features. We observe that most SSC studies in the literature favored the late fusion as mentioned in \cite{roldao20223d}. The study \cite{dourado2022data} employed early and late fusions simultaneously, while the study in \cite{lin2023multi} incorporating multi-scale feature fusion. In this research we selected the optimal approach based on performance metrics, including uncertainty quantification represented by standard deviations. We summarise our contributions as follows:
\begin{itemize}
\vspace{-0.1cm}
\item We propose a hybrid architecture with dual heads to simultaneously learn from multiple data modalities of a single scene, leveraging a combined loss function with a re-weighting method. This design improves learning in scenarios with intra-class diversity and inter-class ambiguity by incorporating loss from 2D RGB semantics to the 3D geometries by F-TSDF.
\item We enhance the overall results by implementing the ITRM block with a hyperbolic tangent activation function applied to identity features. This approach optimises the learning process by emphasizing positive signals for visible spaces and negative signals for occluded regions, ensuring compatibility with the characteristics of F-TSDF data.
\item We evaluate different RGB semantics fusion strategies by incorporating model performance uncertainty. Using K-fold cross-validation, we compute the average scores along with their corresponding standard deviations. This comprehensive analysis facilitates the selection of fusion methods that effectively validate the model's generalisation across diverse scenarios.
\end{itemize}
\vspace{-0.2cm}
\section{Method} \label{sec:Method}
\begin{figure}[tb]
  \centering
  \includegraphics[height=3.5cm]{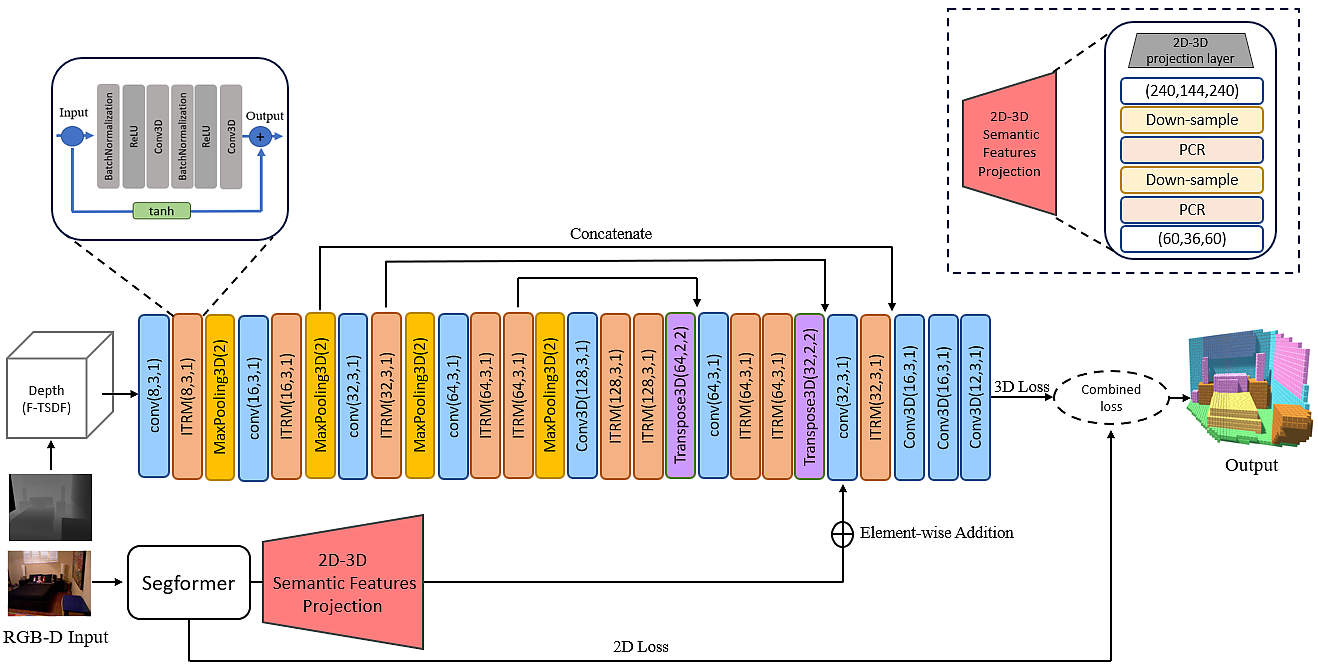}
  \caption{MDBNet is a dual-head network that processes 2D RGB semantics via a pre-trained Segformer with 2D-3D projection (include PCR blocks) and geometric data via a 3D CNN with ITRM blocks. The network optimises a combined loss, which is a weighted sum of 3D loss and 2D semantics loss.}
  \label{fig:MDBNet}
\end{figure}
\vspace{-0.1cm}
\subsection{Overall Framework}
The architecture of the proposed MDBNet is depicted in Figure \ref{fig:MDBNet}. This model features a dual-head network, facilitating learning simultaneously from each network head within a single pipeline. The system processes each scene using two distinct modalities: a 2D input consisting of RGB image at a resolution of 640$\times$480, and depth map data preprocessed as the form of F-TSDF for data representation within 3D space, which captures geometric information with dimensions of 240$\times$144$\times$240. We leverage the Segformer, a pre-trained transformer model for image semantic segmentation, to extract the 2D semantic features, which are subsequently projected into 3D space. For the 3D input, we adopt the foundational structure of the 3D U-Net CNN, as utilized in \cite{dourado2021edgenet}, with a custom adaptation of the residual block. This adaptation includes adding Tanh on identity features. The model generates an output with a four-dimensional structure sized 60$\times$36$\times$60$\times$12. The 12 channels represent the dataset classes ranging from 0 to 11. Class 0 is designated for empty spaces, whereas the remaining classes represent various object categories found in the NYUv2 \cite{silberman2012indoor} and NYUCAD \cite{firman2016structured} datasets, including ceiling, floor, wall, window, chair, bed, sofa, table, TV, furniture, and objects. Further details on this architecture will be discussed in the subsequent subsections.
\vspace{-0.1cm}
\subsection{2D Semantic Features }
\vspace{-0.2cm}
The incorporation of 2D RGB semantic features is motivated by their ability to enhance intra-class consistency and inter-class distinction within the SSC problem. Specifically, RGB semantics add surface features to the objects in scenes, features that are absent in methods relying solely on depth maps as input. Transfer learning emerges as the most effective strategy for this adaptation process. It facilitates the efficient extraction of these RGB semantic features, enabling the system to benefit from learning more diverse features across larger dataset. Consequently, to optimize RGB input utilization, we employed the Segformer `B5' model, which is known for its superior accuracy and performance \cite{NEURIPS2021_64f1f27b}. This Segformer model pre-trained on ImageNet and fine-tuned on the ADE20K dataset at a resolution of 640$\times$640, leverages high-resolution image processing, aligning closely with the resolution of images in the NYU datasets \cite{silberman2012indoor,firman2016structured}. Given the limited size of the NYU dataset and its class overlap with ADE20K, it presents an ideal scenario for transfer learning. We adopted a transfer learning strategy by keeping the encoder's weights fixed and initializing the decoder's weights with those pre-trained on ADE20K, followed by fine-tuning on the NYU datasets \cite{wang2023semantic}.

\subsection{2D-3D Features Projection}
\vspace{-0.1cm}
Features extracted from 2D RGB images are projected and mapped onto the corresponding coordinates in 3D space by taking the advantage of the existed depth map input. Aligned with the projection method described in \cite{liu2018see}, we utilized the depth values from the depth image $I_{depth}$, along with the intrinsic camera matrix \( K \in \mathbb{R}^{3 \times 3} \) and the extrinsic camera matrix \( [R|t] \in \mathbb{R}^{3 \times 4} \) to project a pixel \( p_{u,v} \) represented in homogeneous coordinates as \( [u, v, 1]^T \) from the 2D image plane to a 3D point \( p_{x,y,z} \), also in homogeneous coordinates \( [X, Y, Z, 1]^T \). This projection is accomplished using the camera projection equation referenced as Equation \ref{eq:cam_proj}:
\vspace{-0.1cm}
\begin{equation}
		\begin{aligned}
			p_{u,v} = K [R|t] p_{x,y,z},
			\label{eq:cam_proj}
		\end{aligned}
  \end{equation}
to map the 2D features into scene surfaces in the 3D space. Then, these volumetric surface features are fused with the F-TSDF input within 3D network branch.

\subsection{3D Features Fusion Strategies} \label{sec:3DFeaturesFusionStrategies}
\vspace{-0.1cm}
Different fusion methods based on element-wise addition were implemented to assess the model's performance, including early, mid, and late fusions. The aim of investigating different fusion methods is to identify the best location to add the projected RGB semantic features into the geometric information represented by F-TSDF. Early fusion involved combining the full-resolution projected 3D surface features 240$\times$144$\times$240 with the F-TSDF input prior to their introduction into the 3D network branch. For mid and late fusions, the projected 3D surface features were downsampled to align with the resolutions of the network's intermediate 15$\times$9$\times$15 and later 60$\times$36$\times$60 layers, respectively. This downsampling process employed the Planar Convolution Residual (PCR) block \cite{li2023front}, a variant of the Dimensional Decomposition Residual (DDR) block \cite{li2019rgbd}, which breaks down the standard 3D convolution into three sequential one-dimensional layers along three orthogonal axes. The PCR uses planar convolutions with kernel dimensions where one of the three sizes is 1, preserving the planar characteristics of the 3D scene and reducing parameter count relative to standard residual blocks.

\subsection{Identity Transformed within full pre-activation Residual Module (ITRM)}
\vspace{-0.1cm}
We propose a modification to the residual blocks by incorporating a hyperbolic tangent (Tanh) function on the identity features. The Tanh activation function has been employed in various research contexts, particularly in scenarios where TSDF or SDF are used as input. Its primary purpose in such cases is to manage data distributions within a normalized range, aligning with the inherent data range of TSDF or SDF, as demonstrated in \cite{Park2019DeepSDFLC, Weder2020NeuralFusionOD}. In the domain of SSC, the Tanh activation function has been applied to part of identity features, albeit in a different context \cite{zhang2019cascaded}. Our research extends this exploration by investigating additional context for the application of Tanh, aiming to enhance its integration within residual blocks. The residual blocks in our model adopt the full pre-activation design outlined in \cite{he2016identity}, where batch normalization (BN) and the rectified linear activation function (ReLU) are applied before the convolution layers in a reverse order compared to the standard design, in which BN and ReLU are applied after the convolution layers. This reversal order facilitates smoother information propagation and performance optimisation. The Equations \ref{eq:bn_re} and \ref{eq:ident}:
\vspace{-0.1cm}
\begin{equation}
		\begin{aligned}
			x' = f({BN}(x_l)),
			\label{eq:bn_re}
		\end{aligned}
  \end{equation}  
\begin{equation}
		\begin{aligned}
			x_{l+1} = x_l \:+ F(x',W_l),
			\label{eq:ident}
		\end{aligned}
  \end{equation}
 
illustrate the relationship between the input and output of the full pre-activated residual block, where the input to the $l-th$ residual block is $x_l$ and the output is $x_{l+1}$. The function $f$ represents the activation function applied to the normalized input $x_l$. The residual function $F(x',W_l)$ represents, for example, a series of two convolutional layers, each with a 3$\times$3 filter, applied to $x'$, the pre-activated input in Equation \ref{eq:bn_re}. The term $W_l$ includes a collection of weights (including biases) associated with the $l-th$ residual block. We have modified the full pre-activation residual block design by applying a non-linear transformation with the Tanh function on the identity $x_l$, as illustrated in Equation \ref{eq:tanh_ident}:
\vspace{-0.1cm}
\begin{equation}
		\begin{aligned}
			x_{l+1} = Tanh(x_l) \:+ F(x',W_l).
			\label{eq:tanh_ident}
		\end{aligned}
  \end{equation}
In the F-TSDF representation, voxels in visible or empty spaces above surfaces are given values ranging from 0 to 1, while those in occluded areas have values from -1 to 0, creating steep gradients at objects surfaces \cite{song2017semantic}. The application of the Tanh function is particularly advantageous in this context, as it preserves the sign of the input with positive signals for visible space and negative ones for occluded regions, while normalizing the values to a range between [-1, 1]. This property is crucial for distinguishing between occluded areas and visible surfaces. Additionally, it optimises the learning process by ensuring that the data within the network compatible with the nature of positive and negative values within F-TSDF data, leading to more stable learning. 
\subsection{Combined Loss Function}
We supervise the two inputs of MDBNet jointly using a combined loss function that merges the 2D semantic loss and the 3D loss for SSC, employing a weighted sum approach. This method utilizes a weighting parameter \( \lambda \) to balance the contributions of the two losses, designated as $L_{SS}$ for 2D semantic loss and $L_{SSC}$ for the 3D SSC loss. The combined loss function is formulated in the following Equation \ref{eq:comb_loss}:

\vspace{-0.1cm}
\begin{equation}
		\begin{aligned}
			L =  \lambda L_{SS}\:+ L_{SSC}.
			\label{eq:comb_loss}
		\end{aligned}
	\end{equation}
Aligned with \cite{wang2023semantic}, we employed the smooth cross-entropy loss, denoted as $L_{SS}$, to measure the loss for 2D RGB semantic predictions. The $L_{SSC}$ weighted cross entropy loss \cite{alawadh20243d}, evaluates the model's performance in 3D space, specifically using F-TSDF after integrating projected 2D semantic features in our context. $L_{SSC}$ combines the benefits of resampling and class-sensitive learning to address the inherent class imbalance in the data. It employed a smoothed weights through an unsupervised clustering algorithm, K-means. The computation of $L_{SSC}$ loss assesses the discrepancy between the predicted label $p$ and the genuine label $y$ across the voxels of a scene $A$. For each voxel $v$ within $A$, the predicted and actual labels for a given voxel $v$ are indicated by $p_v$ and $y_v$, respectively. Each voxel label is assigned a specific weight $w_v$ using the reweighing method based on K-means clustering. The loss function is defined as follows in Equation \ref{eq:WCE}:
\vspace{-0.1cm}
\begin{equation}
		\begin{aligned}
			L_{SSC}(p,y) = - \sum\limits_{v=1}^A w_v \:.\:  y_v \:.\: log p_v.
			\label{eq:WCE}
		\end{aligned}
	\end{equation}

\section{Implementation Details} \label{ImplementationDetails}
The implementation of this work is divided into three main phases: preprocessing, training and validation, and evaluation. The code can be accessed at: [URL is hidden for the double blind review].
\subsection{Data Preparation}
\vspace{-0.1cm}
We encoded all 2D depth maps from the NYUv2 and NYUCAD datasets into 3D space using F-TSDF. The processed data is saved and can be used across multiple designs. 
We aligned the 3D scenes with Manhattan world assumption, which is related to the direction of gravity. The defined 3D space dimensions are 4.8 meters in width, 2.88 meters in height, and 4.8 meters in depth. With a voxel grid size of 0.02 meters, this configuration results in a volumetric resolution of 240$\times$144$\times$240 voxels. The TSDF truncation value is set to 0.24 meters, optimizing the balance between detail capture and computational efficiency. Both data resampling and correspondence between 3D spatial points and 2D RGB pixels using depth maps are established in this stage. 

\subsection{Training and Validation}
\vspace{-0.1cm}
We conducted our experiments using the PyTorch framework, on a single Nvidia RTX 8000 GPU. 
Both 2D and 3D network branches are trained simultaneously with MDBNet. Due to the two types of input representation —the 2D RGB and the 3D geometrical input represented by F-TSDF—we employed different learning rates to achieve effective performance as demonstrated in \cite{yao2022modality}. Additionally, we adopted different schedulers and optimisers fitted to our network branches contexts. For the 2D input modality (RGB), we employed a pre-trained Segformer `B5' model, which was fine-tuned on the ADE20K dataset at an image resolution of 640$\times$640. The model weights were downloaded from Hugging Face \cite{segformer_b5_finetuned_ade_640_640}. In the pre-trained model, we kept the encoder's weights fixed and fine-tuned the decoder layers, starting with a learning rate of \(1 \times 10^{-4}\). Following the approach suggested by \cite{wang2023semantic}, we used the AdamW optimizer with 0.05 weight decay, and learning rate governed by a cosine decay policy, starting from the initial value and decreasing to a minimum of \(1 \times 10^{-7}\). For the 3D input modality, we opted for mini-batch Stochastic Gradient Descent (SGD) with a momentum of 0.9 and a weight decay of \(5 \times 10^{-4}\). The OneCycleLR scheduler was utilized to adjust the learning rate, beginning at 0.01. We trained the MDBNet model for 100 epochs, with batch sizes set to 4 for training and 2 for validation. To mitigate the risk of overfitting on the training dataset, we incorporated an early stopping as a regularization method \cite{moradi2020survey} with a patience setting of 15 epochs. In our loss function, we experimented with a coefficient \( \lambda \) set to 1 and normalized the scale of \( L_{SS} \) to match that of \( L_{SSC} \) by setting \( \lambda \) to 0.5. The model exhibited stability across both configurations and demonstrated effective learning. Although the score ranges for both settings showed considerable overlap, a slightly higher SSC score was observed with \( \lambda = 1 \), achieving 60.1 \( \pm \) 1.0 compared to 59.2 \( \pm \) 1.3 with \( \lambda = 0.5 \).
Furthermore, to ensure the performance reliability of our results, we implemented K-fold cross-validation, dividing the training set into three folds at random, and preserving the weights from each fold for subsequent evaluation on the test set, thereby quantifying the model's performance uncertainty.
\vspace{-0.1cm}
\section{Evaluation: Methods and Variations} \label{sec:Evaluation}
\vspace{-0.1cm}
\subsection{Datasets}
\vspace{-0.1cm}
Our research leverages the NYUv2 and NYUCAD datasets as benchmarks for conducting our experiments. NYUv2 consists of 1449 realistic RGB-D indoor scenes captured via a Kinect sensor with a resolution of 640$\times$480.
The datasets divided into 795 training instances and 654 testing instances. However, as discussed in \cite{song2017semantic}, there is some misalignment between the depth images and the corresponding 3D labels in the NYUv2 dataset, which makes it difficult to evaluate accurately. To address this problem, we use the high-quality NYUCAD synthetic dataset, which projects depth maps from ground truth annotations and avoids misalignment. 
\subsection{Metrics}
\vspace{-0.1cm}
We adopt Precision, Recall, and IoU as the evaluation measures for the SSC, following the approach of Song et al. \cite{song2017semantic}. For the semantic scene completion task, both the observed surface and occluded regions are evaluated. We present the mIoU scores for each semantic class, excluding the empty class. In the scene completion task, all non-empty voxels are classified as `1', while empty voxels are labeled as `0'. The binary IoU is computed for the occluded regions in the view frustum along with precision and recall measures. We have observed that there's no standardized method for selecting the scene completion area, leading to slight variations among researchers in the field. Some researchers, as seen in \cite{liu2018see} select the occupied occluded voxels while the empty occluded voxels are re-sampled. On the other hand, SPAwN \cite{dourado2022data} bypasses re-sampling step for empty occluded voxels and evaluates all unoccupied voxels. Other studies, such as PALNet \cite{li2019depth}, DDRNet \cite{li2019rgbd}, and AICNet \cite{li2020anisotropic}, include all occupied voxels in the scene, combining visible surfaces with occluded regions for scene completion evaluation. In this research, we adopted the approach outlined in \cite{liu2018see}, evaluating all occluded occupied voxels and re-sampling empty occluded ones. As highlighted in \cite{liu20242d, li2020anisotropic}, the mIoU metric is considered more critical than IoU. Nonetheless, the results for all metrics were averaged across K-fold cross-validation to derive the final scores.
\section{Experiments} \label{sec:Experiments}
\subsection{Ablation Study} 
\vspace{-0.1cm}
In this section, we conduct ablation studies on the NYUCAD dataset to evaluate the effectiveness of our proposed RGB feature fusion methods and the various components of our model design.
\paragraph{Fusion Strategies} The model with the proposed combined loss function only was trained using various methods to fuse the 3D projected RGB semantic features as explained in Section \ref{sec:3DFeaturesFusionStrategies}. The results, as reflected in the average scores presented in Table \ref{tab:f_abl}, indicate that our model is capable of learning effectively using these different fusion strategies. Among them, the late fusion method demonstrated the best averaged score. 
Specifically, we observed that the TV object was not well recognized in some folds when using the early and middle fusion methods, whereas it was consistently recognized across all folds with the late fusion approach. Consequently, we selected the late fusion approach for RGB semantic features to further evaluate the model's performance across different components.
\begin{table}[tb]
\centering
  \caption{Ablation studies using different RGB features fusion methods.}
  \label{tab:f_abl}
  {
  \begin{tabular}{@{}l@{\hspace{20pt}}l@{\hspace{20pt}}l@{}}
    \toprule
    Fusion Method & SC-IoU\% & SSC-mIoU\%\\
    \midrule
    Early  &  80.5 &  57.1\\
    Middle &  79.3 &  55.8\\
    Late   &  79.3 &  59.0\\ 
    \bottomrule
  \end{tabular}
  }
\end{table}

\begin{table}[tb]
 \centering
  \caption{Ablation studies on the NYUCAD dataset evaluating MDBNet components with RGB-D input.}
  \label{tab:c_abl}
  {
  \begin{tabular}{c@{\hspace{3pt}}@{\hspace{3pt}}c@{\hspace{3pt}}c}
    \hline
   Method & SC-IoU\% & SSC-mIoU\% \\
    \hline
   $L_{ss} + L_{SSC}$ (re-weighting)  & 79.3 &  59.0 \\
    $L_{ss} + L_{SSC}$ (re-sampling)  & 80.5&  52.5\\   
    $L_{ss} + L_{SSC}$ (re-weighting) + ITRM  & 79.8 &  60.1 \\ 
    \hline
  \end{tabular}} 
\end{table}


\paragraph{Architecture Components.} To confirm the impact of each component within our MDBNet, we modified the model from \cite{alawadh20243d} by integrating new components and conducted comprehensive experiments to evaluate their contributions, as detailed in Table \ref{tab:c_abl}. Initially, we trained our model with RGB-D input and applied our combined loss, which includes the re-weighting 3D loss \cite{alawadh20243d}, achieving an SSC score of 59.0\%. In the second experiment, we replaced the re-weighting loss with a resampling-based loss from \cite{song2017semantic}. This substitution resulted in a significant decrease of 6.5 percentage points (pp) in the SSC score, underlining the critical role of both RGB features and our combined loss in the model's performance. In the third experiment, we enhanced the 3D branch of MDBNet by replacing the original residual blocks with the proposed ITRM blocks. This enhancement yielded further improvements, achieving an SSC score of 60.1\%, a 7.6 pp increase compared to the second experiment's score of 52.5\%.



\subsection{Comparison with State-of-the-Art Methods}
\vspace{-0.1cm}
\begin{table*}
    \centering
    \caption{Results on the NYUv2 dataset include averages and standard deviations for Precision, Recall, IoU, and mIoU metrics. The `*' represents the view-volume architecture type.} 
    \label{NYUv2_table} 
  \scalebox{0.75}{
    \begin{tabular}{c c c c c c c c c c c c c c c c c c}
        \hline
        \multirow{2}{*}{Method} & 
        \multirow{2}{*}{Input} & 
        \multirow{2}{*}{Res.} &  
        \multicolumn{3}{c}{Scene Completion (SC)} & 
        \multicolumn{12}{c}{Semantic Scene Completion (SSC)}\\
      \cline{4-18}
  &&& Prec. & Recall & IoU & ceil. & floor & wall & win. & chair & bed & sofa & table & tvs & furn. & objs & mIoU \\
\hline 
AMMNet$_{Segformer}$\cite{wang2024unleashing}& RGB-D & (60,60) & 90.5 & 82.1 & 75.6 & 46.7 & 94.2 & 43.9 & 30.6 & 39.1 & 60.3  & 54.8 & 35.7 & 44.4 & 48.2 & 35.3  & 48.5\\
CleanerS\cite{wang2023semantic}& RGB-D & (60,60) & 88.0 & 83.5 & 75.0 & 46.3 & 93.9 & 43.2 & 33.7 & 38.5 & 62.2  & 54.8 & 33.7 & 39.2 & 45.7 & 33.8  & 47.7\\
SISNet(voxel)\cite{cai2021semantic}& RGB-D & (60,60) & 87.6 & 78.9 & 71.0 & 46.9 & 93.3 & 41.3 & 26.7 & 30.8 & 58.4 & 49.5 & 27.2 & 22.1 & 42.2 & 28.7  & 42.5 \\
PCANet*\cite{li2023front}& RGB-D & (240,60) & 89.5 & 87.5 & 78.9 & 44.3 & 94.5 & 50.1 & 30.7 & 41.8 & 68.5  & 56.4 & 32.6 & 29.9 & 53.6 & 35.4  & 48.9\\
SPAwN\cite{dourado2022data}& RGB-D & (240,60) & 82.3 & 77.2 & 66.2 & 41.5 & 94.3 & 38.2 & 30.3 & 41.0 & 70.6  & 57.7 & 29.7 & 40.9 & 49.2 & 34.6  & 48.0\\
\hline
MDBNet (Ours) & RGB-D & (240,60) & 80.3$\pm 3.7$ & 81.8 $\pm 6.5$ & 67.6$\pm 2.1$ & 47.2 & 92.6 & 49.9 & 47.6 & 46.8 & 66.2 & 62.1 & 37.1 & 35.7 & 45.2 & 36.9 & 51.6$\pm 1.5$\\
\hline
\end{tabular}}
\end{table*}
 \begin{table*}   
    \centering
   \caption{Results on the NYUCAD dataset include averages and standard deviations for Precision, Recall, IoU, and mIoU metrics. The `*' represents the view-volume architecture type.}
    \label{NYUCAD_table} 
  \scalebox{0.75}{
    \begin{tabular}{c c c c c c c c c c c c c c c c c c}
        \hline
        \multirow{2}{*}{Method} & 
        \multirow{2}{*}{Input} & 
        \multirow{2}{*}{Res.} &  
        \multicolumn{3}{c}{Scene Completion (SC)} & 
        \multicolumn{12}{c}{Semantic Scene Completion (SSC)}\\
      \cline{4-18}
  &&&{Prec.} &{Recall} &{IoU} &{ceil.} & {floor} & {wall} &{win.} & {chair} & {bed} &{sofa}  & {table} & {tvs} & {furn.} &{objs} & {mIoU} \\
 \hline 
AMMNet$_{Segformer}$\cite{wang2024unleashing}& RGB-D & (60,60) & 92.4 & 88.4 & 82.4 & 61.3 & 94.7 & 65.0 & 38.9 & 58.1 & 76.3  & 73.2 & 47.3 & 46.6 & 62.0 & 42.6  & 60.5\\
SISNet(voxel)\cite{cai2021semantic}& RGB-D & (60,60) & 92.3 & 89.0 & 82.8 & 61.5 & 94.2 & 62.7 & 38.0 & 48.1 & 69.5  & 59.3 & 40.1 & 25.8 & 54.6 & 35.3  & 53.6\\

SPAwN\cite{dourado2022data}& RGB-D & (240,60) & 84.5 & 87.8 & 75.6 & 65.3 & 94.7 & 61.9 & 36.9 & 69.6 & 82.2  & 72.8 & 49.1 & 43.6 & 63.4 & 44.4  & 62.2\\
PCANet*\cite{li2023front}& RGB-D & (240,60) & 92.1 & 84.3 & 86.3 & 54.8 & 93.1 & 62.8 & 44.3 & 52.3 & 75.6  & 70.2 & 46.9 & 44.8 & 65.3 & 45.8  & 59.6\\
\hline
MDBNet (Ours) & RGB-D & (240,60) & 85.0$\pm 1.7$
 & 93.0 $\pm 1.2$ & 79.8$\pm 0.8$ & 67.4 & 93.6 & 64.1 & 52.4 & 59.5 & 72.5 & 69.3 & 45.0 & 41.5 & 53.1 & 42.4 & 60.1$\pm 1.0$\\
 \hline
\end{tabular}}
\end{table*}
Experiments were conducted to evaluate the performance of our proposed approach on scene completion and semantic scene completion tasks, using the NYUv2 and NYUCAD datasets. Quantitative comparisons of our MDBNet results with SOTA approaches are detailed in Tables \ref{NYUv2_table} and \ref{NYUCAD_table}. Unlike previous studies, which did not specify the performance uncertainty, we averaged our scores across three folds to more accurately represent generalization performance. Due to the variations in how researchers select the scene completion area, as discussed in Section \ref{sec:Evaluation}, these differences do not necessarily show true performance gaps between SOTA models. Also, \cite{liu20242d, li2020anisotropic} highlighted the importance of mIoU over IoU. However, for a fair comparison, we will focus on semantic scene completion, which related to the object area and are measured using standardized criteria. We compare MDBNet with SOTA methods that utilize hybrid architectures, focusing on voxel-based semantic segmentation on the NYUv2 dataset, as shown in Table \ref{NYUv2_table}. Our approach significantly outperforms current SOTA models, achieving a remarkable increase in mIoU scores by 3.1 pp and 2.7 pp over the previously leading methods, AMMNet$_{Segformer}$ \cite{wang2024unleashing} which employed Segformer pretrained model for 2D RGB features, and PCANet \cite{li2023front}, respectively. This establishes MDBNet as the new benchmark in SOTA performance. The efficacy of MDBNet is further confirmed on the NYUCAD dataset as depicted in Table \ref{NYUCAD_table}. MDBNet shows an increase in the average mIoU scores compared to top previous methods, such as PCANet \cite{li2023front}.  Furthermore, although our design surpasses SPAwN on the NYUv2 dataset, it demonstrates performance comparable to the more resource-intensive SPAwN model, which utilizes semantics priors calculated using surface normals and sequential training for 2D and 3D models. 
\subsection{Qualitative Analysis}
\vspace{-0.1cm}
\begin{figure}[tb]
  \centering
   \includegraphics[height =2.65cm, width=8cm]{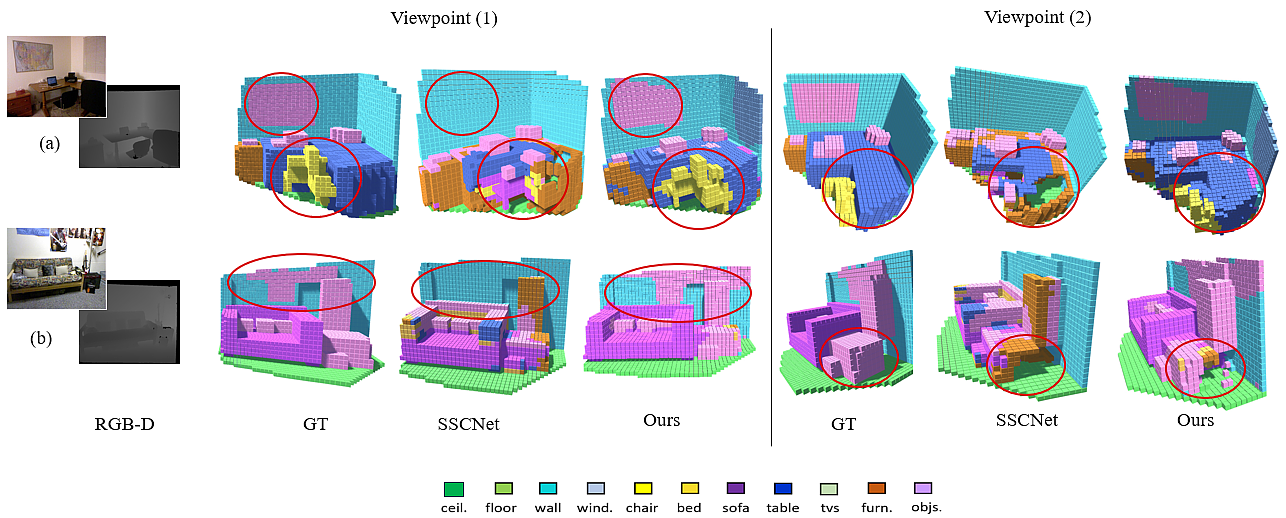}
  \caption{ Comparison of SSC results on the NYUv2 dataset: SSCNet (depth maps) vs. MDBNet (RGB-D). Objects are color-coded, with circles marking key differences between GT and predictions.}
  \label{fig:MDBNet_NYU}
\end{figure}
\begin{figure}[tb]
  \centering
  \includegraphics[height =4cm, width=8.5cm]{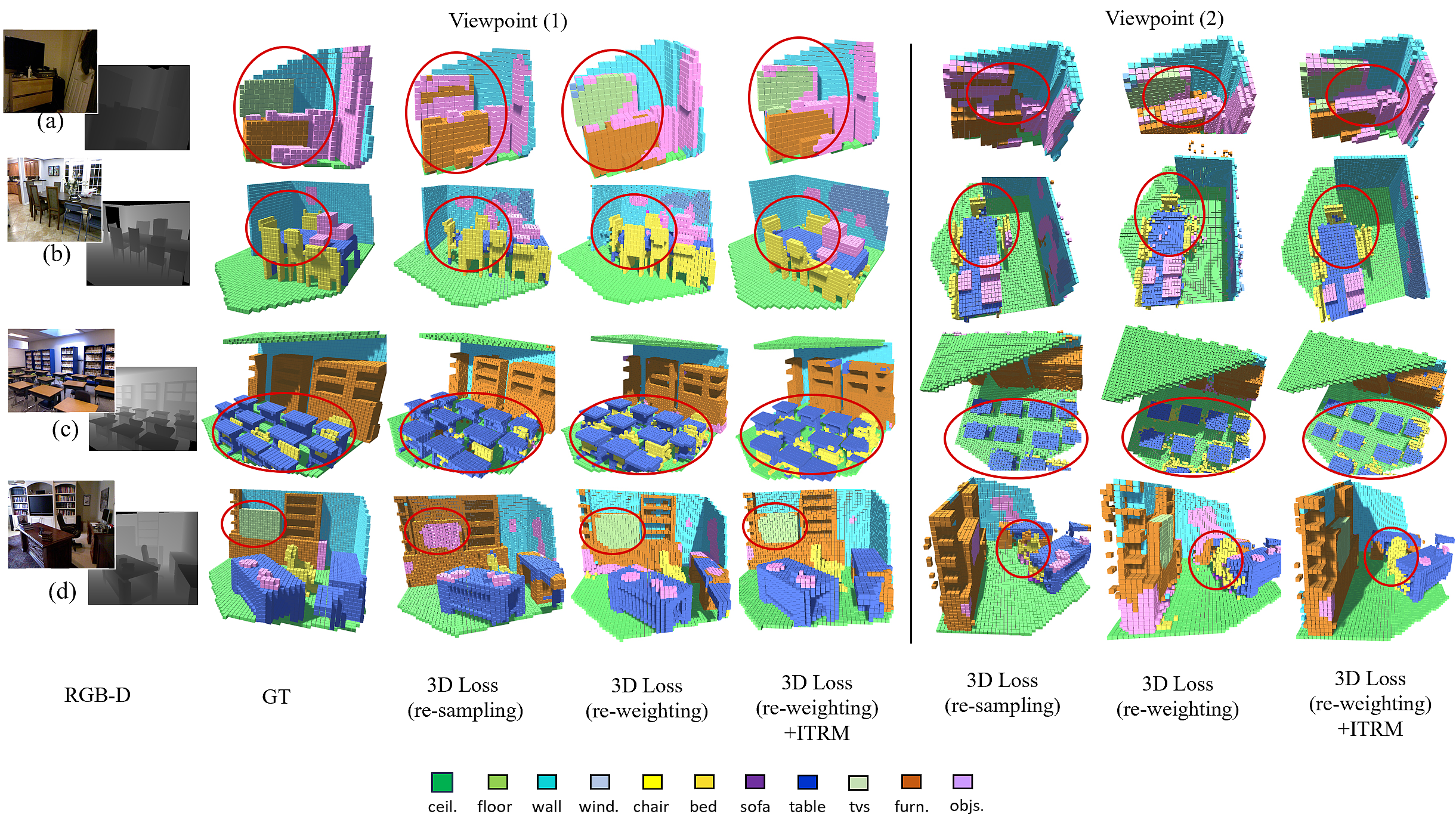}
  \caption{SSC results with different components on NYUCAD dataset. From left to right: (1) RGB-D input; (2) GT; (3) combined loss with re-sampling; (4) combined loss with re-weighting; (5) combined loss (using re-weighting) with ITRM blocks. Objects are color-coded, with circles highlighting key differences between GT and predictions.}
  \label{fig:MDBNet_NYUCAD}
\end{figure}
To highlight the superiority of the MDBNet design and its success in generating more precise predictions, we present a series of visual comparisons using the NYUv2 dataset, as illustrated in Figure \ref{fig:MDBNet_NYU}. These comparisons, made between our method and SSCNet \cite{song2017semantic}, demonstrate the enhanced prediction accuracy offered by our approach. By employing re-weighting method \cite{alawadh20243d} within our combined loss and ITRM, we achieve enhanced scene completion, particularly in the occluded parts of the scenes, as demonstrated in (a) and (b) of Figure \ref{fig:MDBNet_NYU}. Additionally, by extracting semantic features from the RGB inputs, MDBNet exhibits superior performance, even surpassing the ground truth (GT) 3D volumes in certain regions. For instance, in Figure \ref{fig:MDBNet_NYU} in (a), the RGB image shows both object and window existed on the walls. Our model successfully predicts the object and window voxels on the walls where they are absent in the GT 3D volumes. To illustrate the effectiveness of MDBNet's components, Figure \ref{fig:MDBNet_NYUCAD} showcases various scenarios within the NYUCAD dataset, comparing when our combined loss function uses weighting based on re-sampling \cite{song2017semantic} within the 3D loss, when it applies class re-weighting \cite{alawadh20243d}, and when employing re-weighting \cite{alawadh20243d} and incorporating ITRM. The incorporation of class re-weighting in our combined loss significantly enhances the model's ability to identify underrepresented classes, such as TVs and chairs, as shown in Figure \ref{fig:MDBNet_NYUCAD} in (a), (c), and (d). Additionally, our final design MDBNet offers better recognition of chairs with various shapes in the same figure in (b), (c), and (d), and it ensures enhanced differentiation between tables and chairs, as evident in (b) and (c). MDBNet model effectively recognizes challenging classes like windows and TVs, showcasing its robustness and adaptability. Additional results are available on our GitHub.

\section{Conclusion} \label{sec:Conclusion}
\vspace{-0.1cm}
In this study, we addressed the SSC problem, which involves the simultaneous determination of volumetric occupancy and object classification from a single RGB-D input, offering a limited perspective. We tackled key challenges in this area, including the imbalance within 3D spaces of indoor environments, diversity within object classes, and ambiguity among different object classes. MDBNet offers an effective solution by implementing several components, including our combined loss function with ITRM blocks incorporation, the investigation of the RGB fusion placement, and benchmark training methods such as K-fold cross-validation. We demonstrated an improvement in the SSC task on the NYU datasets.

\bibliographystyle{IEEEbib}
\bibliography{main}

\begin{thebibliography}{10}

\bibitem{alawadh2022room}
Mona Alawadh, Yihong Wu, Yuwen Heng, Luca Remaggi, Mahesan Niranjan, and Hansung Kim,
\newblock ``Room acoustic properties estimation from a single 360° photo,''
\newblock in {\em European Signal Processing Conference (EUSIPCO)}, 2022, pp. 857--861.

\bibitem{gao2022review}
Shaohua Gao, Kailun Yang, Hao Shi, Kaiwei Wang, and Jian Bai,
\newblock ``Review on panoramic imaging and its applications in scene understanding,''
\newblock {\em IEEE Transactions on Instrumentation and Measurement}, vol. 71, pp. 1--34, 2022.

\bibitem{liang2021sscnav}
Yiqing Liang, Boyuan Chen, and Shuran Song,
\newblock ``Sscnav: Confidence-aware semantic scene completion for visual semantic navigation,''
\newblock in {\em IEEE international conference on robotics and automation (ICRA)}, 2021, pp. 13194--13200.

\bibitem{kim2022immersive}
Hansung Kim, Luca Remaggi, Aloisio Dourado, Teofilo~de Campos, Philip~JB Jackson, and Adrian Hilton,
\newblock ``Immersive audio-visual scene reproduction using semantic scene reconstruction from 360 cameras,''
\newblock {\em Virtual Reality}, vol. 26, no. 3, pp. 823--838, 2022.

\bibitem{song2017semantic}
Shuran Song, Fisher Yu, Andy Zeng, Angel~X Chang, Manolis Savva, and Thomas Funkhouser,
\newblock ``Semantic scene completion from a single depth image,''
\newblock in {\em CVPR}, 2017, pp. 1746--1754.

\bibitem{roldao20223d}
Luis Roldao, Raoul De~Charette, and Anne Verroust-Blondet,
\newblock ``3d semantic scene completion: a survey,''
\newblock {\em IJCV}, pp. 1--28, 2022.

\bibitem{pan2023understanding}
Yancheng Pan, Fan Xie, and Huijing Zhao,
\newblock ``Understanding the challenges when 3d semantic segmentation faces class imbalanced and ood data,''
\newblock {\em IEEE Transactions on Intelligent Transportation Systems}, vol. 24, no. 7, pp. 6955--6970, 2023.

\bibitem{zhang2019cascaded}
Pingping Zhang, Wei Liu, Yinjie Lei, Huchuan Lu, and Xiaoyun Yang,
\newblock ``Cascaded context pyramid for full-resolution 3d semantic scene completion,''
\newblock in {\em ICCV}, 2019, pp. 7801--7810.

\bibitem{li2019depth}
Jie Li, Yu~Liu, Xia Yuan, Chunxia Zhao, Roland Siegwart, Ian Reid, and Cesar Cadena,
\newblock ``Depth based semantic scene completion with position importance aware loss,''
\newblock {\em IEEE Robotics and Automation Letters}, vol. 5, no. 1, pp. 219--226, 2019.

\bibitem{tang2022not}
Jiaxiang Tang, Xiaokang Chen, Jingbo Wang, and Gang Zeng,
\newblock ``Not all voxels are equal: Semantic scene completion from the point-voxel perspective,''
\newblock in {\em AAAI}, 2022, pp. 2352--2360.

\bibitem{dourado2021edgenet}
Aloisio Dourado, Teofilo~E De~Campos, Hansung Kim, and Adrian Hilton,
\newblock ``Edgenet: Semantic scene completion from a single rgb-d image,''
\newblock in {\em ICPR}, 2021, pp. 503--510.

\bibitem{dourado2022data}
Aloisio Dourado, Frederico Guth, and Teofilo de~Campos,
\newblock ``Data augmented 3d semantic scene completion with 2d segmentation priors,''
\newblock in {\em IEEE Winter Conference on Applications of Computer Vision (WACV)}, 2022, pp. 3781--3790.

\bibitem{alawadh20243d}
Mona Alawadh, Mahesan Niranjan, and Hansung Kim,
\newblock ``3d semantic scene completion from a depth map with unsupervised learning for semantics prioritisation,''
\newblock in {\em 2024 IEEE International Conference on Image Processing (ICIP)}. IEEE, 2024, pp. 3348--3354.

\bibitem{silberman2012indoor}
Nathan Silberman, Derek Hoiem, Pushmeet Kohli, and Rob Fergus,
\newblock ``Indoor segmentation and support inference from rgbd images,''
\newblock in {\em ECCV}, 2012, pp. 746--760.

\bibitem{firman2016structured}
Michael Firman, Oisin Mac~Aodha, Simon Julier, and Gabriel~J Brostow,
\newblock ``Structured prediction of unobserved voxels from a single depth image,''
\newblock in {\em CVPR}, 2016, pp. 5431--5440.

\bibitem{Park2019DeepSDFLC}
Jeong~Joon Park, Peter~R. Florence, Julian Straub, Richard~A. Newcombe, and S.~Lovegrove,
\newblock ``Deepsdf: Learning continuous signed distance functions for shape representation,''
\newblock {\em 2019 IEEE/CVF Conference on Computer Vision and Pattern Recognition (CVPR)}, pp. 165--174, 2019.

\bibitem{Weder2020NeuralFusionOD}
Silvan Weder, Johannes~L. Sch{\"o}nberger, Marc Pollefeys, and Martin~R. Oswald,
\newblock ``Neuralfusion: Online depth fusion in latent space,''
\newblock {\em 2021 IEEE/CVF Conference on Computer Vision and Pattern Recognition (CVPR)}, pp. 3161--3171, 2020.

\bibitem{lin2023multi}
Di~Lin, Haotian Dong, Enhui Ma, Lubo Wang, and Ping Li,
\newblock ``Multi-head multi-scale feature fusion network for semantic scene completion,''
\newblock in {\em 2023 International Conference on Artificial Intelligence and Education (ICAIE)}. IEEE, 2023, pp. 57--61.

\bibitem{NEURIPS2021_64f1f27b}
Enze Xie, Wenhai Wang, Zhiding Yu, Anima Anandkumar, Jose~M. Alvarez, and Ping Luo,
\newblock ``Segformer: Simple and efficient design for semantic segmentation with transformers,''
\newblock in {\em NeurIPS}, 2021, pp. 12077--12090.

\bibitem{wang2023semantic}
Fengyun Wang, Dong Zhang, Hanwang Zhang, Jinhui Tang, and Qianru Sun,
\newblock ``Semantic scene completion with cleaner self,''
\newblock in {\em CVPR}, 2023, pp. 867--877.

\bibitem{liu2018see}
Shice Liu, Yu~Hu, Yiming Zeng, Qiankun Tang, Beibei Jin, Yinhe Han, and Xiaowei Li,
\newblock ``See and think: Disentangling semantic scene completion,''
\newblock {\em NeurIPS}, vol. 31, 2018.

\bibitem{li2023front}
Jie Li, Qi~Song, Xiaohu Yan, Yongquan Chen, and Rui Huang,
\newblock ``From front to rear: 3d semantic scene completion through planar convolution and attention-based network,''
\newblock {\em IEEE TMM}, 2023.

\bibitem{li2019rgbd}
Jie Li, Yu~Liu, Dong Gong, Qinfeng Shi, Xia Yuan, Chunxia Zhao, and Ian Reid,
\newblock ``Rgbd based dimensional decomposition residual network for 3d semantic scene completion,''
\newblock in {\em CVPR}, 2019, pp. 7693--7702.

\bibitem{he2016identity}
Kaiming He, Xiangyu Zhang, Shaoqing Ren, and Jian Sun,
\newblock ``Identity mappings in deep residual networks,''
\newblock in {\em ECCV}, 2016, pp. 630--645.

\bibitem{yao2022modality}
Yiqun Yao and Rada Mihalcea,
\newblock ``Modality-specific learning rates for effective multimodal additive late-fusion,''
\newblock in {\em The Association for Computational Linguistics (ACL)}, 2022, pp. 1824--1834.

\bibitem{segformer_b5_finetuned_ade_640_640}
{NVIDIA},
\newblock ``Segformer b5 finetuned ade 640x640,'' \url{http://tinyurl.com/segformerb5}, 2024,
\newblock Accessed: 2024-02-06.

\bibitem{moradi2020survey}
Reza Moradi, Reza Berangi, and Behrouz Minaei,
\newblock ``A survey of regularization strategies for deep models,''
\newblock {\em Artificial Intelligence Review}, vol. 53, no. 6, pp. 3947--3986, 2020.

\bibitem{li2020anisotropic}
Jie Li, Kai Han, Peng Wang, Yu~Liu, and Xia Yuan,
\newblock ``Anisotropic convolutional networks for 3d semantic scene completion,''
\newblock in {\em CVPR}, 2020, pp. 3351--3359.

\bibitem{liu20242d}
Xianzhu Liu, Haozhe Xie, Shengping Zhang, Hongxun Yao, Rongrong Ji, Liqiang Nie, and Dacheng Tao,
\newblock ``2d semantic-guided semantic scene completion,''
\newblock {\em International Journal of Computer Vision}, pp. 1--20, 2024.

\bibitem{wang2024unleashing}
Fengyun Wang, Qianru Sun, Dong Zhang, and Jinhui Tang,
\newblock ``Unleashing network potentials for semantic scene completion,''
\newblock in {\em CVPR}, 2024, pp. 10314--10323.

\bibitem{cai2021semantic}
Yingjie Cai, Xuesong Chen, Chao Zhang, Kwan-Yee Lin, Xiaogang Wang, and Hongsheng Li,
\newblock ``Semantic scene completion via integrating instances and scene in-the-loop,''
\newblock in {\em CVPR}, 2021, pp. 324--333.

\end{thebibliography}

\vspace{-0.2cm}
\end{document}